# A Framework to Counteract Suboptimal User-Behaviors in Exploratory Learning Environments: an Application to MOOCs

Sébastien Lallé        Cristina Conati

Department of Computer Science, The University of British Columbia
{lalles, conati}@cs.ubc.ca

## Abstract

While there is evidence that user-adaptive support can greatly enhance the effectiveness of educational systems, designing such support for exploratory learning environments (e.g., simulations) is still challenging due to the open-ended nature of their interaction. In particular, there is little a priori knowledge of which student's behaviors can be detrimental to learning in such environments. To address this problem, we focus on a data-driven user-modeling framework that uses logged interaction data to learn which behavioral or activity patterns should trigger help during interaction with a specific learning environment. This framework has been successfully used to provide adaptive support in interactive learning simulations. Here we present a novel application of this framework we are working on, namely to Massive Open Online Courses (MOOCs), a form of exploratory environment that could greatly benefit from adaptive support due to the large diversity of their users, but typically lack of such adaptation. We describe an experiment aimed at investigating the value of our framework to identify student's behaviors that can justify adapting to, and report some preliminary results.

## Introduction

Providing personalized instruction is important to facilitate human learning, as it is demonstrated by the effectiveness of one-to-one human tutoring and of instruction provided by Intelligent Tutoring Systems, namely learning environments that leverage Artificial Intelligence and Machine Learning to adapt teaching to the needs of individual learners, as good teachers do (Woolf 2008). Delivering effective personalized support to learning activities requires having student models that can identify which learners are in need of support, when support should be given, and the right level and content of assistance. There has been extensive research and positive results for building such models for structured educational activities such as problem solving and question answering, where answer correctness can be clearly defined. Building such models is more challenging for exploratory or open ended educational activities (e.g., interactive simulations, educational games), during which it is difficult to define a priori which students' behaviors or activities are conductive to learning. Yet, learners often need support during these activities, because their open-ended nature can be conducive to confusion, lack of motivation, disorientation.

To design adaptation in exploratory learning environments, we proposed a *Framework for User Modeling and Adaptation* (FUMA) that uses logged interaction data to learn which user behavioral or activity patterns should trigger help (Kardan and Conati 2011; Conati and Kardan 2013). In FUMA, clustering is first applied to identify groups of students who learn similarly with the environment based on their logged data alone. Next, association rule mining is used to extract the distinguishing behavioral patterns of each clustered group of students. These association rules are then used to classify new users and trigger real-time adaptive interventions built upon the underlying behavioral or activity patterns. FUMA was successfully applied in previous work to provide adaptive support to students learning with the CSP applet, an interactive simulation for the AC3 constraint satisfaction algorithm. That work showed that an adaptive version of the CSP applet derived from FUMA generated better learning than its non- adaptive counterpart, and was especially useful for students with lower initial domain knowledge (Kardan and Conati 2015). FUMA was also applied to a more complex simulation for teaching concepts related to electric circuits (CKK). This work showed that FUMA can derive students models similar in accuracy and performance to the one in Kardan and Conati (2015), and can identify student behaviors to drive adaptive interventions (Fratamico et al. 2017).

In this paper, we first describe FUMA, and then discuss a new application we are working on, namely using FUMA to capture and counteract suboptimal behaviors of users learn-ing from Massive Online Open Courses (MOOCs). Although the popularity of MOOCs has risen to engage hundred thousands of students in higher-education, they currently suffer from a key limitation typical of many on-line learning environments: lack of personalization. MOOCS can greatly benefit from personalization, due to the highly varied demographics of their users, but work on providing intelligent student-adaptive support in MOOCS is still limited (Pardos et al. 2013).

We have started to explore the value of using FUMA to provide this intelligent support by focusing on edX data related to students' video usage (e.g., pausing, fast forwarding…), to identify video watching behaviors that can affect learning performance. The long-term goal is to leverage these behaviors to design and deliver adaptation in edX MOOCs aimed at promoting efficient video behaviors. We focus on students' video watching behaviors because videos typically account for a significant amount of the learning material in MOOCS, and thus are often described as the MOOCS "backbone" (Yousef et al. 2014). However, not all students use and benefit from videos in the same way (Kim et al. 2014; Guo et al. 2014), calling for the need of adaptive support that can promote effective video watching behaviors. To the best of our knowledge, our research is the first to explore the feasibility of recognizing and delivering such adaptation in a data-driven way.

In the rest of this paper, we first describe FUMA. Next we describe the MOOC dataset on which we apply FUMA, and the experiment we envision to deliver adaptation in MOOCs. Then we present preliminary results obtained with FUMA on the dataset, related to the unsupervised clustering of the students based on their video watching behaviors. Lastly, we report previous work and conclude.

## FUMA Framework

The overall structure of FUMA is shown in Fig. 1, and consists of two main phases: *Behavior Discovery* and *User Classification*.

### Behavior Discovery

This phase aims at identifying the behaviors that could be relevant for driving adaptation. First, each user's interaction data is pre-processed into normalized feature vectors that summarize the relevant elements of the user's behavior in the MOOC, one vector per student. Students are then clustered using these vectors so as to identify groups of learners with similar interaction behaviors. Clustering is done via the Genetic Algorithm K-means (GA K-means) (Krishna and Murty 1999), with the optimal number of cluster $k$ selected based on three well-established measures to do so: C-index, Calinski-Harabasz, and Silhouettes (Wang et al. 2009). K-means was chosen because it is a common and computationally simple clustering algorithm, and it was found to perform equally well as other clustering approaches (Kardan and Conati 2011). However, because K-means is subject to converging to local maxima, FUMA addresses this problem by using the genetic version of K-means (Krishna and Murty 1999). GA K-means does an evolutionary search with many different starting centroids to minimize the chance that it converges to a local maximum. The resulting clusters are used to label the student's data (i.e., each student is assigned to one cluster).

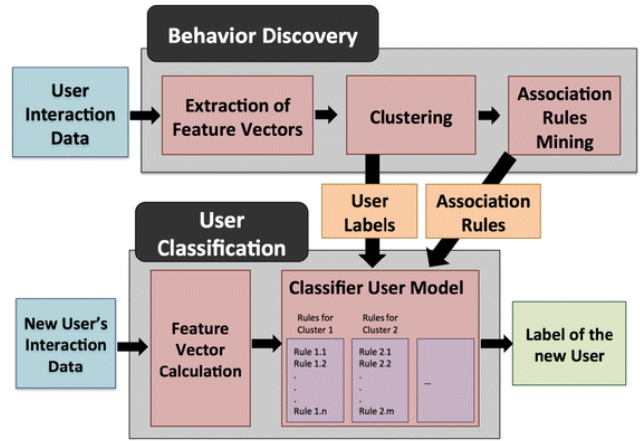

*Fig. 1. FUMA's workflow.*

When an appropriate measure of learning performance is available, it can be used to compare and qualify the clusters. FUMA carries-out this comparison by using a one-way ANOVA with the given performance measure as the dependent measure, and the clusters as factor.

The next step is to identify the distinctive interaction behaviors in each cluster via association rule mining using Weka's Hotspot algorithm (Holmes et al. 1994). This process generates association rules for each cluster in the form of a tree and extracts the common behavior patterns in terms of class association rules in the form of X ➔ c, where X is a set of behavior-value pairs and c is the predicted class label for the data points where X applies.

In addition to being used for online classification (see next subsection), these rules also provides insights on the reasons for performing poorly and well in the course. Such information could, in turn, translate into practical suggestions on how to adapt the MOOC. For instance, the MOOC could recommend the student to not fast-forward if this behavior is associated to poor learning performance and is exhibited too frequently by the student.

### Rule-Based User Classification

In this phase, the labeled clusters and the corresponding association rules extracted in Behavior Discovery are used for classification (see Classifier User Model box in Fig. 1). Specifically, as a new student interacts with the system, FUMA evaluates which association rules are satisfied by the student's behaviors, and assigns the student to the cluster corresponding to the satisfied rules. The main difficulty with this approach is that the student's behaviors might match rules pertaining to different clusters. In that case, simply classifying the student into the cluster with the highest number of matched rules fails to account for the fact that some rules have better support than others (e.g., apply to more students in the corresponding cluster). FUMA addresses this issue by computing a membership score $S_A$ for a given cluster *A* as follow:

$$S_A = \frac{\sum_{i=1}^{m} T_{ri} \times C_{ri}}{\sum_{i=1}^{m} C_{ri}}$$

Where $r_i$ is the *ith* rules in cluster *A* (out of a total of *m* rules), $C_{ri}$ is the confidence of $r_i$ (confidence for rule "X ➔ c" is the probability that *c* occurs when *X* does), and $T_{ri}$ indicates whether the rule $r_i$ has been satisfied by the student ("0" if no, "1" if yes). It should be noticed that this formula takes into account all the rules in a given cluster rather than just the satisfied ones. The rationale behind this choice is that, in FUMA the rules that do not apply to the student are also important for determining the final label. For instance, it is important to penalize the score of a cluster *A* if the major rules in *A* (i.e., those with a high confidence) are not satisfied by the student. The student is ultimately assigned to the cluster with the highest $S_A$ score.

In addition to classifying students, this phase serves the purpose of delivering real-time adaptation. Specifically, FUMA returns the specific association rules describing the student's behaviors that caused the classification. These behaviors can then trigger real-time interventions designed to encourage productive behaviors and discourage detrimental ones, as done in (Kardan and Conati 2015). Thus, the strength of FUMA is that it supports the entire adaptation process in a data-driven way, from discovering and modeling relevant behaviors of the students to delivering real-time adaptation.

### Previous Evaluations of FUMA

FUMA has been applied in two interactive learning simulations, the CSP and CCK Applets, as mentioned in the Introduction. Results showed that the behaviors identified by FUMA can classify unseen students as "low" or "high" learners significantly better than chance, with accuracies up to 86% for the CSP Applet (Conati and Kardan 2013) and 85% for the CKK Applet (Fratamico et al. 2017). FUMA was also found to outperform standard classifiers (e.g., random forests, decision trees, SVM…) at distinguishing high versus low learners in Conati and Kardan (2013).

Kardan and Conati (2015) formally evaluated the effectiveness of the adaptive support provided by FUMA in the CSP Applet. Specifically, FUMA was used to deliver in real-time prompts aimed at correcting inefficient behaviors exhibited by students classified as low learners during the interaction with the CSP applet. The prompts recommended behaviors previously identified by FUMA at the "Behavior Discovery" phase to be associate with high learners. The results of this study showed that the version of the CSP applet with adaptive support improved students learning, compared to the CSP applet with no support or with support generated at random.

## Experiment Set-Up

### Dataset

We use the data collected from the Summer 2014's session of the Computer Science 101 (CS101) MOOC deployed at the Stanford's University[1]. This course comprises of 33 videos and 29 quizzes and exercises spanned over 6 weeks. 141,988 students enrolled in this course, out of whom 16,994 (11.8%) passed (final grade of 0.8 or above). Fig. 2 shows the number of student who were active (i.e., logged-in and performed at least one action) at each week. This figure shows a typical drop-out rate in MOOC, with a very high amount of students who dropped at the very first week, usually because these students enrolled to try and see if they are interested in the course.

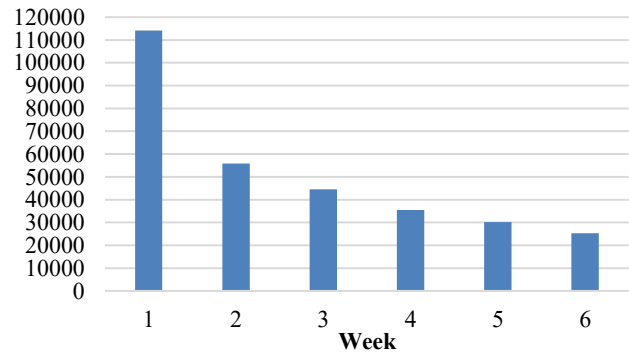

*Fig. 2. Number of active students at each week in the course.*

Each action performed by the students is recorded by edX. In our case, we focus on video logs, which include what videos the student loaded, as well as the sequence of actions performed on each video. The available actions are play, pause, seek, change the speed of the video, and stop. The logs also include information about the time in the video

---

[1] lagunita.stanford.edu/courses/Engineering/CS101/Summer2014/about

when the action was performed, as well as the new time after seeking.

### Features generated from MOOC's data

To mine relevant student video-watching behaviors, FUMA requires a set of features that capture important aspects of the student interaction with the videos ("Extraction of Feature Vectors" on Fig. 1). For this experiment we generated a battery of 21 features for each student, listed in Table 1.

14 of these features have been linked in previous work to course completion, performance on quiz and learning, and thus are good candidates for FUMA to explore behaviors that can explain learning outcome and course completion, e.g., (Athira et al. 2015; Liu et al. 2017; Liu and Xiu 2017; Wen and Rosé 2014; Li et al. 2016). These features are: frequency of actions, number of video watched, mean weekly video coverage, proportion of videos rewatched, proportion of interrupted videos, the average duration of pauses and time spent speed-up videos.

In addition, we measure the average seek length[2] and the average number of time the student rewatched videos. Seeking can be used in different ways, e.g., to skip parts of a video or come back and rewatch them, thus the average seek length can qualify better how the students use the *seek* action while watching videos. The number of time the videos are rewatched can reveal how engaged are the students in the course material. Furthermore, we leverage information related to the standard deviation of the weekly video coverage, number of rewatch per videos, duration of pauses, duration of seek jumps, and time speeding-up. We use features based on standard deviation because they can provide insights about the consistency of some of the behaviors exhibited by the students (e.g., does the student consistently take long pauses, or a mixture of long and short pauses).

---

- **Frequency of each type of video action** performed (play, pause, seek backward, seek forward, change speed, stop)
- **Frequency** and **number** of all actions performed
- **Total number of video watched**
- **Proportion of rewatched videos**
- **Average** and **SD of the number of time a video was rewatched**
- **Proportion of interrupted videos**
- **Average** and **SD Weekly video coverage** (prop. of video watched each week)
- **Average** and **SD duration of video pauses**
- **Average** and **SD seek length**
- **Average** and **SD time speeding-up the video**

*Table 1: List of features used for running FUMA.*

---

[2] Backward seek lengths are encoded as negative values here.

### Application of FUMA

To account for the fact that the number of students change as weeks elapse due to students dropping-out, we run FUMA on the student's data at Week 2, Week 3 and Week 4, for a total of 3 runs. We ignored Week 1 at this step because a large majority of the students already drop at this week (see Fig. 2) and may provide irrelevant behaviors for FUMA. We also did not ran FUMA on Week 5 and 6 because we aim at delivering adaptive assistance early in the course, when students are likely to need it the most.

The set of features in Table 1 are computed at each of the 3 weeks we study, and FUMA is run on these features using a nested 10-folds cross-validation approach. Specifically, students are split into 10 folds, with each fold being used in turn for testing while the remaining 9 folds are used for training the rule-based classifiers. The performance of the classifiers are then averaged over the 10 testing folds. Nested cross-validation is applied within the training sets to set FUMA's parameters while avoiding contaminating the test fold. These parameters include for instance defining the optimal number of clusters and the thresholds for the minimum required rule support.

## Preliminary Results

We present here preliminary results pertaining to the first phase of FUMA, namely the clustering analysis, to understand whether FUMA can infer relevant clusters based on the students' video watching activity. This preliminary analysis aims at informing us about the potential value of FUMA, as identifying relevant clusters is necessary for building the rule-based classifiers (see section above).

The optimal number of clusters identified by FUMA was 2 for all weeks included in the analysis (Week 2-4). To assess the quality of the clusters, we measure how large is the difference among the clusters in terms of three measures commonly used in MOOCs to assess learning and achievement: 1. the *final grade* obtained by the students at the end of the course (defined here as the average of the scores obtained on all quizzes); 2. the proportion of students who passed the course; 3. the drop-out rate. We formally compare the clusters by conducting, at each of the three weeks we study, a statistical analysis with these three measures as dependant variables. Because these measures are of different nature, we ran different statistical models: ANOVA for the final grade, and multinomial logistic regressions for the proportion of students who passed and the drop-out rate. For each dependent variable, we ran the appropriate model with the cluster (2 levels) as the factor, and adjusted the p-values for family-wise error by applying the Holm-Bonferroni based on the number of comparisons made.

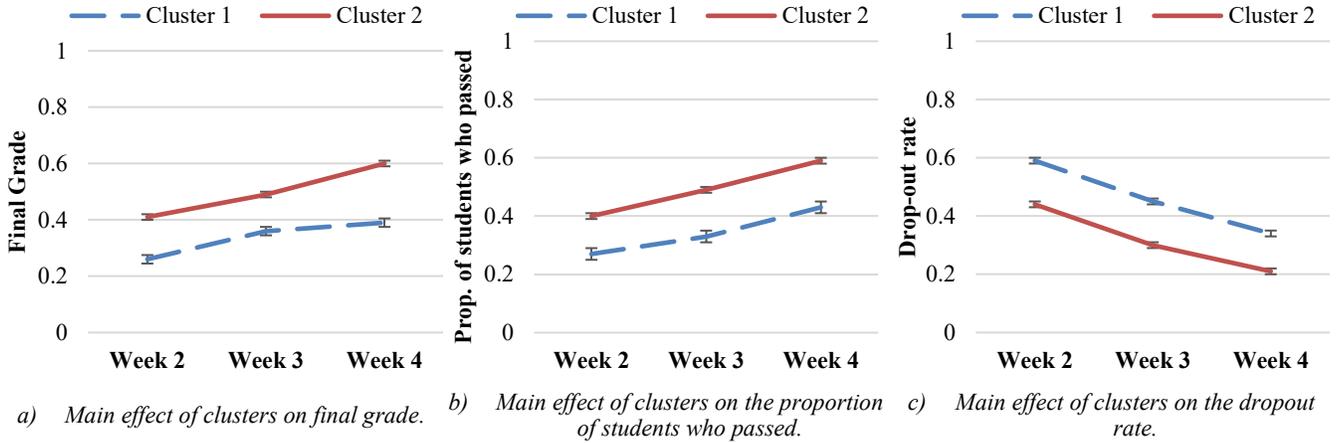

a) *Main effect of clusters on final grade.*
b) *Main effect of clusters on the proportion of students who passed.*
c) *Main effect of clusters on the dropout rate.*

*Fig. 3. Charts showing the evolution of the three studied dependent variables on average across weeks in each cluster.*

As a result we found a significant main effect of *cluster* at each week. The *p*-values and effect sizes for these effects are reported in Table 2. We report effect sizes as large for $\eta^2 > 0.26$, medium for $\eta^2 > 0.13$, and small otherwise. Fig. 3 shows in addition the evolution across weeks of each dependant variable on average in both clusters.

Overall, these results reveal significant differences in terms of students' performance across cluster in all cases, indicating that video watching behaviors can significantly impact performance. In particular, students in cluster-2 were found to outperform students in cluster-1 in terms of final grade and proportion of students who passed, while exhibiting a significantly lower drop-out rate. This means that the video-watching behaviors exhibited by students in cluster-2 can lead to better performance and less drop-outs, while it is the opposite for students in cluster-1. These findings indicate that FUMA can identify relevant behaviors, and thus it is justified to apply the next steps of FUMA on these data to identify these specific behaviors and understand whether the can drive adaptive assistance, our long-term goal.

| Dep. variable | Week 2 | Week 3 | Week 4 |
|---|---|---|---|
| *Final grade* | p < .0001<br>$\eta^2 = 0.13$ | p < .0001<br>$\eta^2 = 0.14$ | p < .0001<br>$\eta^2 = 0.17$ |
| *Prop. passed* | p < .0001<br>$\eta^2 = 0.14$ | p < .0001<br>$\eta^2 = 0.14$ | p < .0001<br>$\eta^2 = 0.15$ |
| *Drop-out rate* | p < .001<br>$\eta^2 = 0.11$ | p < .001<br>$\eta^2 = 0.09$ | p < .001<br>$\eta^2 = 0.08$ |

*Table 2: Results of the statistical analysis of the effect of clusters on the three studied dependent variables.*

Table 2 shows that the effect sizes remain constant across week, indicating that the differences among the cluster do not greatly vary overtime. This finding is interesting for our goal of delivering adaptive assistance, because it suggests that video watching behaviors can be informative already at Week 2, i.e., quite early in the course, when the assistance is likely to matter the most. We can also notice that the performance of the students increase on average across week, which is not surprising as the students who remains active in the later weeks are more likely to be the motivated ones. This trend, however, indicate that the specific behaviors exhibited by the students at each week might change, which can be revealed by the association rules generated by FUMA, i.e., the next step in the FUMA's workflow after the clustering analysis.

To understand which behaviors were the most useful to split the students, we extracted the top 5 video features that are the most discriminant to cluster the students, as shown in Table 3.

| Week | Top 5 features | Dir. |
|---|---|---|
| Week 2 | -Number of watched videos* | + |
| | -Prop. of rewatched videos* | + |
| | -Mean weekly video coverage* | + |
| | -Mean video pauses* | + |
| | -Freq. of all actions* | + |
| Week 3 | - Prop. of interrupted videos* | + |
| | - Mean weekly video coverage* | + |
| | - Mean video pauses* | + |
| | - Frequency of video speed change* | - |
| | - Mean weekly video coverage* | + |
| Week 4 | - Mean weekly video coverage* | + |
| | - Prop. of rewatched videos* | + |
| | - Prop. of interrupted videos* | - |
| | - Frequency of actions* | + |
| | - Mean video pauses* | + |

*Table 3: Top 5 most discriminant features at each week (ordered from top to bottom). * indicates a statistical difference among the cluster. A positive directionality (dir.) indicates that the features is higher in cluster-2 than in cluster-1.*

We statistically compared these features by running, for each of them, a Wilcoxon pairwise comparison test with the cluster as the factor (significant effects at reported in Table 3). Overall, the nature and directionality of the features in Table 3, as well as their statistical comparisons, indicate that the students in Cluster-2 were more engaged in the course (higher proportion of video watched and rewatched, lower proportion of videos interrupted) and more active during video watching (more frequent actions, longer pauses) than students in Cluster-1. This indicates that the clusters are based on the level of engagement and activity of the students during video watching, which in turns can explain the differences in performance of students among clusters.

While the aforementioned results reveal that students in cluster 2 were more engaged during video watching and performed significantly better than in cluster 1, we can notice that there are still a quite large proportion of students who did not pass in cluster 2 (from 60% at Week-2 to 42% at Week-4, Fig. 3) or even dropped, while some students in cluster 1 were able to pass (from 27% at Week-2 to 44% at Week-4). The fact that many students passed in cluster-1 while being not as active might indicate that these students had high prior knowledge and thus did not need to spend too much time on the videos, or that they were watching some of the videos offline. We are planning to explore this second idea, as students who watched videos offline should be discarded from the analysis due to the fact that parts of their data are missing. As for the students who failed in cluster-2, a closer look at their data reveal that they skipped some or most of the quizzes, while remaining engaged in the course. This might indicate that their intent was not to obtain the certification, but just to follow the course. Overall, these findings suggest that information about quizzes (number of attempt…) could compliment the video watching features used in our analysis with FUMA, in order to also split students based on their intent in the course (e.g., complete the quizzes to get certified, or ignore the quizzes and the certification).

The next phase of FUMA (Rule-Based Classification) will aim at extracting the most salient patterns of behaviors in each cluster, to inform the design of adaptive support. One approach for such support would be, for instance, recommending to students in cluster-1, some of the relevant behaviors exhibiting by the students in cluster-2 who achieve high learning outcomes.

## Related Work

There has been extensive work on analyzing student clickstream data in MOOCs, both offline and online. Offline studies typically relies on data-mining techniques to identify relevant behaviors and learning strategies. In particular, several of these studies have leveraged the same data-mining techniques used in FUMA: clustering and association rule mining. For instance, Athira et al. (2015), Liu et al. (2017), Liu and Xiu (2017) and Wen and Rosé (2014) have mined clusters and association rules to identify relationships among student's behaviors and their engagement in the course. In Srilekshmi et al. (2016) and Wen and Rosé (2014), they mine association rules to identify behaviors that can be predictive of course completion. Boroujeni and Dillenbourg (2018) identified the different strategies used the students to explore the learning content by mining clusters and patterns from interaction data. While in the above work they leverage descriptive measures related to overall usages of the course resources (e.g., number of quizzes taken, video watched or posts on the forum…), in our work we focus on fine-grained behaviors related to the videos, by exploring the rich video logs tracked in edX. Furthermore, unlike in Athira et al. (2015), Liu et al. (2017), Boroujeni and Dillenbourg (2018), Liu and Xiu (2017) and Wen and Rosé (2014), FUMA uses the identified clusters and associations rules not only offline, but also online to identify ineffective learners and behaviors during interaction with the MOOC. Only Srilekshmi et al. (2016) used association rules online, to predict course completion, however, in that work each association rule was evaluated individually, and without comparison against a baseline, whereas FUMA can aggregate the predictions made by a set of several relevant association rules (see above, "Rule-Based User Classification" section).

Other works analyzed students data online to model relevant student traits and behaviors during interaction with a MOOC. Most of these works focused on building machine learning models to predict course completion from log data, e.g., Li et al. (2016), Nagrecha et al. (2017), and Zhao et al. (2016), or the students' levels of mastery of the learning content (Wang et al. 2016; Pardos et al. 2013). However these models were not aimed at informing the design of adaptive interventions as FUMA does.

A few studies have evaluated the value of personalization in MOOCs. Rosen et al. (2017) and Ketamo (2014) provided recommendations on what resources to explore next in the course based on the students' past visited pages and current knowledge level. While in Rosen et al. (2017) the tested recommendations were not successful, in Ketamo (2014) they were found to improve learning outcome. Brinton et al. (2015) found that hiding too-advanced parts of the course to low learners significantly improves their engagement. Sonwalkar (2013) showed that adapting the course content to the student's learning style increased their engagement in the course. Davis et al. (2017) showed that providing less-active students with recommendation on how to increase participation in the curse improved the course completion rate. While these works have shown that adaptation is MOOCs can be valuable, the range of the tested adaptive mechanisms is still limited. By extracting from data behaviors relevant for adaptation, FUMA could reveal further forms of adaptations that can be valuable in MOOCs, as well as insights on when to deliver them. FUMA can also ease the evaluation of adaptive mechanisms and the replication of these evaluations across MOCCs, as the framework provides functionalities to classify students in real time and deliver relevant forms of adaptation during interaction with the MOOC.

Beyond MOOCs, work in the field of Intelligent Tutoring Systems (ITS) exists on recognizing from data student behaviors and states that can justifying adapting to. For example, there has been work on generating hints from data based on observed students errors during problem solving (Barnes et al. 2008; Shute and Zapata-Rivera 2007; Stamper et al. 2010). Other work leveraged students' data to learn when it is suitable to deliver specific types of help (Razzaq and Heffernan 2006; Verbert et al. 2011; Lallé et al. 2013). While ITSs typically feature structured forms of instruction (e.g., multi-step problem solving), FUMA can be applied to more exploratory, open-ended environments where students can explore the learning material at will.

## Conclusion and Future Work

We presented in this paper the *Framework for User Modeling and Adaptation* (FUMA), that uses students' logged interaction data to design and drive adaptive support in exploratory learning environments. FUMA uses machine learning and data mining to identify from the logged data behaviors or activities that can be detrimental for learning, indicating the students' need for help or their usage of suboptimal learning strategies. Specifically, clustering is applied to identify groups of students who learn similarly. Next, association rule mining is used to extract the distinguishing interaction behaviors of each clustered group of students. FUMA can also recognize these behaviors in real-time and react accordingly, for example by prompting the use of efficient learning strategies.

We described an experiment we are conducting to apply this framework to MOOC data, with the long-term goal of providing personalized support in MOOCs to enhance students' performance and engagement. Such study would be greatly beneficial to MOOCs, as MOOCs currently suffer from a key limitation typical of many online learning environments: a lack of personalization. The need for personalization is particularly crucial in MOOCs due to the highly varied demographics of their users. Preliminary results show that FUMA can identify clusters from MOOC's video watching data with significant differences among the clusters in terms of students' learning outcome and activity in the course.

In future work, we are planning to investigate not only the clustering part of FUMA, but the entire FUMA process, namely to infer distinctive association rules and leverage these rules to classify new users into the appropriate cluster in real time. The association rules will also provide insights about which patterns of behaviors can predict low learning performance, so that adaptation could be delivered to counteract those patterns. Ultimately, we will explore the value of FUMA to deliver such adaptation in real time in MOOCs.

## Acknowledgments

This work is funded by the Data Science Institute of the University of British Columbia.

## References

Aleven, V., Sewall, J., Andres, J.M., Sottilare, R., Long, R., and Baker, R. 2018. Towards Adapting to Learners at Scale: Integrating MOOC and Intelligent Tutoring Frameworks. In *Proceedings of the Fifth Annual ACM Conference on Learning at Scale*, 14. ACM.

Athira, L., Kumar, A., and Bijlani, K. 2015. Discovering Learning Models in MOOCs Using Empirical Data. In *Emerging Research in Computing, Information, Communication and Applications*, 551–567. Springer.

Barnes, T., Stamper, J., Lehman, L., and Croy, M. 2008. A Pilot Study on Logic Proof Tutoring Using Hints Generated from Historical Student Data. In *Proceedings of the Conference on Educational Data Mining*, 197.

Boroujeni, M.S., and Dillenbourg, P. 2018. Discovery and Temporal Analysis of Latent Study Patterns in MOOC Interaction Sequences. In *Proceedings of the 8th International Conference on Learning Analytics and Knowledge*, 206–215. ACM.

Brinton, C.G., Rill, R., Ha, S., Chiang, M., Smith, R., and Ju, W. 2015. "Individualization for Education at Scale: MIIC Design and Preliminary Evaluation. *IEEE Transactions on Learning Technologies* 8 (1): 136–148.

Conati, C., and Kardan, S. 2013. Student Modeling: Supporting Personalized Instruction, from Problem Solving to Exploratory Open Ended Activities. *AI Magazine* 34 (3): 13–26.

Davis, D, Jivet, I., Kizilcec, R.F., Chen, G., Hauff, C., and Houben, G.-J. 2017. Follow the Successful Crowd: Raising MOOC Completion Rates through Social Comparison at Scale. In *Proceedings of the Seventh International Learning Analytics & Knowledge Conference*, 454–463. ACM.

Fratamico, L., Conati, C., Kardan, S., and Roll, I. 2017. Applying a Framework for Student Modeling in Exploratory Learning Environments: Comparing Data Representation Granularity to Handle Environment Complexity. *International Journal of Artificial Intelligence in Education* 27 (2): 320–352.

Guo, P.J., Kim, J., and Rubin, R. 2014. How Video Production Affects Student Engagement: An Empirical Study of MOOC Videos. In *Proceedings of the First ACM Conference on Learning @ Scale Conference*, 41–50. ACM.

Holmes, G., Donkin, A., and Witten, I.H. 1994. Weka: A Machine Learning Workbench. In *Proceedings of the Second Australian and New Zealand Conference on Intelligent Information Systems*, 357–361.

Kardan, S., and Conati, C. 2011. A Framework for Capturing Distinguishing User Interaction Behaviors in Novel Interfaces. In *Proceedings of the Conference on Educational Data Mining*, 159–168.

Kardan, S., and Conati, C. 2015. Providing Adaptive Support in an Interactive Simulation for Learning: An Experimental Evaluation. In *Proceedings of the 33rd Annual ACM Conference on Human Factors in Computing Systems*, 3671–3680. ACM.

Ketamo, H. 2014. Learning Fingerprint: Adaptive Tutoring for MOOCs. In *Proceedings of EdMedia: World Conference on Educational Media and Technology*.

Kim, J, Guo, P.J., Seaton, D.T., Mitros, P., Gajos, K.Z., and Miller, R.C. 2014. Understanding In-Video Dropouts and Interaction Peaks Inonline Lecture Videos. In *Proceedings of the First ACM Conference on Learning @ Scale Conference*, 31–40. ACM.

Krishna, K., and Narasimha Murty, M. 1999. Genetic K-Means Algorithm. *IEEE Transactions on Systems, Man, and Cybernetics, Part B (Cybernetics)* 29 (3): 433–439.

Lallé, S., Mostow, J., Luengo, V., and Guin, N. 2013. Comparing Student Models in Different Formalisms by Predicting Their Impact on Help Success. In *Proceedings of the 16th International Conference on Artificial Intelligence in Education*, 161–170. Springer.

Li, W., Gao, M., Li, H., Xiong, Q., Wen, J., and Wu, Z. 2016. Dropout Prediction in MOOCs Using Behavior Features and Multi-View Semi-Supervised Learning. In *Proceedings of the International Joint Conference on Neural Networks*, 3130–3137. IEEE.

Liu, S., Hu, Z., Peng, X., Liu, Z., Cheng, H., and Sun, J. 2017. Mining Learning Behavioral Patterns of Students by Sequence Analysis in Cloud Classroom. *International Journal of Distance Education Technologies* 15 (1): 15–27.

Liu, T., and Xiu, L.I. 2017. Finding out Reasons for Low Completion in MOOC Environment: An Explicable Approach Using Hybrid Data Mining Methods. *DEStech Transactions on Social Science, Education and Human Science*.

Nagrecha, S., Dillon, J.Z., and Chawla, N.V. 2017. MOOC Dropout Prediction: Lessons Learned from Making Pipelines Interpretable. In *Proceedings of the 26th International Conference on World Wide Web Companion*, 351–359.

Pardos, Z.A., Bergner, Y., Seaton, D.T., and Pritchard, D.E. 2013. Adapting Bayesian Knowledge Tracing to a Massive Open Online Course in edX. In *Proceedings of the Conference on Educational Data Mining*, 137–144.

Razzaq, L., and Heffernan, N.T. 2006. Scaffolding vs. Hints in the Assistment System. In *Proceedings of the International Conference on Intelligent Tutoring Systems*, 635–44. Springer.

Rosen, Y., Rushkin, I., Ang, A., Federicks, C., Tingley, D., and Blink, M.J. 2017. Designing Adaptive Assessments in MOOCs. In *Proceedings of the Fourth ACM Conference on Learning@ Scale*, 233–236. ACM.

Shute, V.J., and Zapata-Rivera, D. 2007. Adaptive Technologies. In *Handbook of Research on Educational Communications and Technology*, Springer, 277–294.

Sonwalkar, N. 2013. The First Adaptive MOOC: A Case Study on Pedagogy Framework and Scalable Cloud Architecture—Part I. *MOOCs Forum* 1: 22–29.

Srilekshmi, M., Sindhumol, S., Chatterjee, S., and Bijlani, K. 2016. Learning Analytics to Identify Students At-Risk in MOOCs. In *Proceedings of the IEEE Eighth International Conference on Technology for Education*, 194–199. IEEE.

Stamper, J., Barnes, T., and Croy, M. 2010. Enhancing the Automatic Generation of Hints with Expert Seeding. In *Proceedings of the International Conference on Intelligent Tutoring Systems*, 31–40. Springer.

Verbert, K., Drachsler, H., Manouselis, N., Wolpers, M., Vuorikari, R., and Duval, E. 2011. Dataset-Driven Research for Improving Recommender Systems for Learning. In *Proceedings of the 1st International Conference on Learning Analytics and Knowledge*, 44–53. ACM.

Wang, K., Wang, B., and Peng, L. 2009. CVAP: Validation for Cluster Analyses. *Data Science Journal* 8: 88–93.

Wang, Z., Zhu, J., Li, X., Hu, Z., and Zhang, M. 2016. Structured Knowledge Tracing Models for Student Assessment on Coursera. In *Proceedings of the Third ACM Conference on Learning@ Scale*, 209–212. ACM.

Wen, M., and Rosé, C.P. 2014. Identifying Latent Study Habits by Mining Learner Behavior Patterns in Massive Open Online Courses. In *Proceedings of the 23rd ACM International Conference on Conference on Information and Knowledge Management*, 1983–1986. ACM.

Woolf, B.P. 2008. *Building Intelligent Interactive Tutors: Student-Centered Strategies for Revolutionizing E-Learning*. Morgan Kaufmann.

Yousef, A.M.F, Chatti, M.A., Schroeder, U., and Wosnitza, M. 2014. What Drives a Successful MOOC? An Empirical Examination of Criteria to Assure Design Quality of MOOCs. In *Proceedings of the IEEE 14th International Conference on Advanced Learning Technologies*, 44–48. IEEE.

Zhao, C., Yang, J., Liang, J., and Li, C. 2016. Discover Learning Behavior Patterns to Predict Certification. In *Proceedings of the 11th International Conference on Computer Science & Education*, 69–73. IEEE.